\documentclass[10pt,twocolumn,letterpaper]{article}

\usepackage{cvpr}              % To produce the CAMERA-READY version

\usepackage{graphicx}
\usepackage{times}
\usepackage{helvet}
\usepackage{courier}
\usepackage{amsmath}
\usepackage{algorithm}
\usepackage{algorithmic}
\usepackage{csquotes}
\usepackage{color}
\usepackage{paralist}
\usepackage{amssymb}
\usepackage{indentfirst}
\usepackage{float}
\usepackage{multirow}
\usepackage{cite}
\usepackage{mathrsfs}

\usepackage{mathrsfs}
\usepackage[colorlinks, linkcolor=red,  anchorcolor=blue, citecolor=blue]{hyperref}
\usepackage{textcomp,booktabs}
\usepackage{amssymb}% http://ctan.org/pkg/amssymb
\usepackage{pifont}% http://ctan.org/pkg/pifont
\usepackage{colortbl}
\definecolor{mygray}{gray}{.9}
\definecolor{mypink}{rgb}{.99,.91,.95}
\definecolor{mycyan}{cmyk}{.3,0,0,0}
\usepackage[capitalize]{cleveref}
\crefname{section}{Sec.}{Secs.}
\Crefname{section}{Section}{Sections}
\Crefname{table}{Table}{Tables}
\crefname{table}{Tab.}{Tabs.}

\def\cvprPaperID{3849} % *** Enter the CVPR Paper ID here
\def\confName{CVPR}
\def\confYear{2023}

\begin{document}

\title{Rethinking Batch Sample Relationships for Data Representation: A Batch-Graph Transformer based Approach}

\author{Xixi Wang, Bo Jiang\thanks{Corresponding author}, ~Xiao Wang, and Bin Luo \\
\emph{School of Computer Science and Technology, Anhui University, Hefei, China} \\
 {\tt\small sissiw0409@foxmail.com}, {\tt\small jiangbo@ahu.edu.cn}, {\tt\small xiaowang@ahu.edu.cn}, {\tt\small ahu\_lb@163.com}
}

% \author{Xixi Wang\\
% %Institution1 address\\
% {\tt\small sissiw0409@foxmail.com}
% \and
% Bo Jiang\thanks{Corresponding author}\\
% {\tt\small jiangbo@ahu.edu.cn}
% \and
% Xiao Wang \\
% {\tt\small xiaowang@ahu.edu.cn}
% \and
% Bin Luo \\
% {\tt\small luobin@ahu.edu.cn} \\
% Anhui University
% }

\maketitle

%%%%%%%%% ABSTRACT
\begin{abstract}
Exploring sample relationships within each mini-batch has shown great potential for learning image representations.
Existing works generally adopt the regular Transformer to model the visual content relationships, ignoring the cues of semantic/label correlations between samples.
Also, they generally adopt the `full' self-attention mechanism which are obviously redundant and also sensitive to the noisy samples.
To overcome these issues, in this paper, we design a simple yet flexible Batch-Graph Transformer (BGFormer) for mini-batch sample representations by deeply capturing the relationships of image samples from both visual and semantic perspectives.
 BGFormer has three main aspects.
 (1) It employs a flexible graph model, termed \textit{Batch Graph} to jointly encode the visual and semantic  relationships of samples within each mini-batch.
(2) It explores the neighborhood relationships of samples by borrowing the idea of sparse graph representation which thus performs robustly, w.r.t., noisy samples.
(3) It devises a novel Transformer architecture that mainly adopts \textit{dual} structure-constrained self-attention (SSA), together with graph normalization, FFN, etc, to carefully exploit the batch graph information for  sample tokens (nodes) representations.
As an application, we apply BGFormer to the metric learning tasks.
Extensive experiments on four popular datasets demonstrate the effectiveness of the proposed model.
\end{abstract}

\section{Introduction}
Learning robust and discriminative image feature representation is a fundamental  task in computer vision applications, such as image classification~\cite{he2016Resnet,dosovitskiy2020ViT}, visual recognition~\cite{jiang2021ph,he2021transreid}, object tracking~\cite{chen2021transformer,fu2022sparsett}, metric learning~\cite{suh2019SM,ermolov2022Hyp} and so on.
With the development of deep learning, Convolutional Neural Networks (CNNs) have been widely used for image  representation learning.
However, CNNs just model the local information of images and thus fail to fully capture the global context information of images.
It is well known that Transformers~\cite{vaswani2017attention,liu2021swin}  have achieved great successes in many visual representation and learning tasks due to their core self-attention mechanism to capture the long-range dependencies among visual units, \textit{e.g.}, patches, regions, etc.
In recent years, various Transformer variants~\cite{han2022vision,touvron2021DeiT,caron2021DINO,hou2022batchformer} have been studied in computer vision task to facilitate the better image feature representation learning problem.
For example, Dosovitskiy et al.~\cite{dosovitskiy2020ViT} develop a Visual Transformer (ViT) architecture to model the contextual relationships of image regions and thus obtain the competitive image classification results.
Liu et al.~\cite{liu2021swin} propose a hierarchical vision Transformer with shifted windows, namely Swin Transformer, to produce the hierarchical feature representation by using hierarchical architecture.
Rao et al.~\cite{rao2021dynamicvit} propose a DynamicViT model to obtain the better classification results by dynamically pruning redundant image patches/regions.
Chen et al.~\cite{chen2021transformer} introduce a Transformer Tracking (TransT) to combine template and search region features by using attention for object tracking task.
Han et al.~\cite{han2022vision} develop a Vision GNN (ViG) architecture to generate graph-level feature by viewing the patches of image as nodes for visual tasks.
%  These pioneering works are based on image patch/region, which just model the context information of intra-image and ignore the correlation of inter-image.

Beyond exploring local patch/region relationships for image representation, recent studies also propose to leverage Transformer to explore sample-level correlations for image   representation learning~\cite{wang2022few,hou2022batchformer,hou2022batchformerv2}.
One representative work is
BatchFormer~\cite{hou2022batchformer,hou2022batchformerv2} which aims to capture the interaction between all samples in each mini-batch to learn robust data representations.
Wang et al.~\cite{wang2022few} recently propose QSFormer to deeply explore the interations between query and support samples for few-shot classification problem.
However, existing models generally directly adopt the ideas of traditional Transformers~\cite{vaswani2017attention,dosovitskiy2020ViT}. Also, they generally only focus on modeling the visual relationships among images, ignoring the important cues of semantic/label correlations among images.
Moreover, existing BatchFormer generally adopts regular self-attention (SA) mechanism which
learns each image
by aggregating the information from all other samples. Obviously, this mechanism is redundant and also sensitive to the noisy samples.

\begin{figure}
\centering
\includegraphics[width=0.45\textwidth]{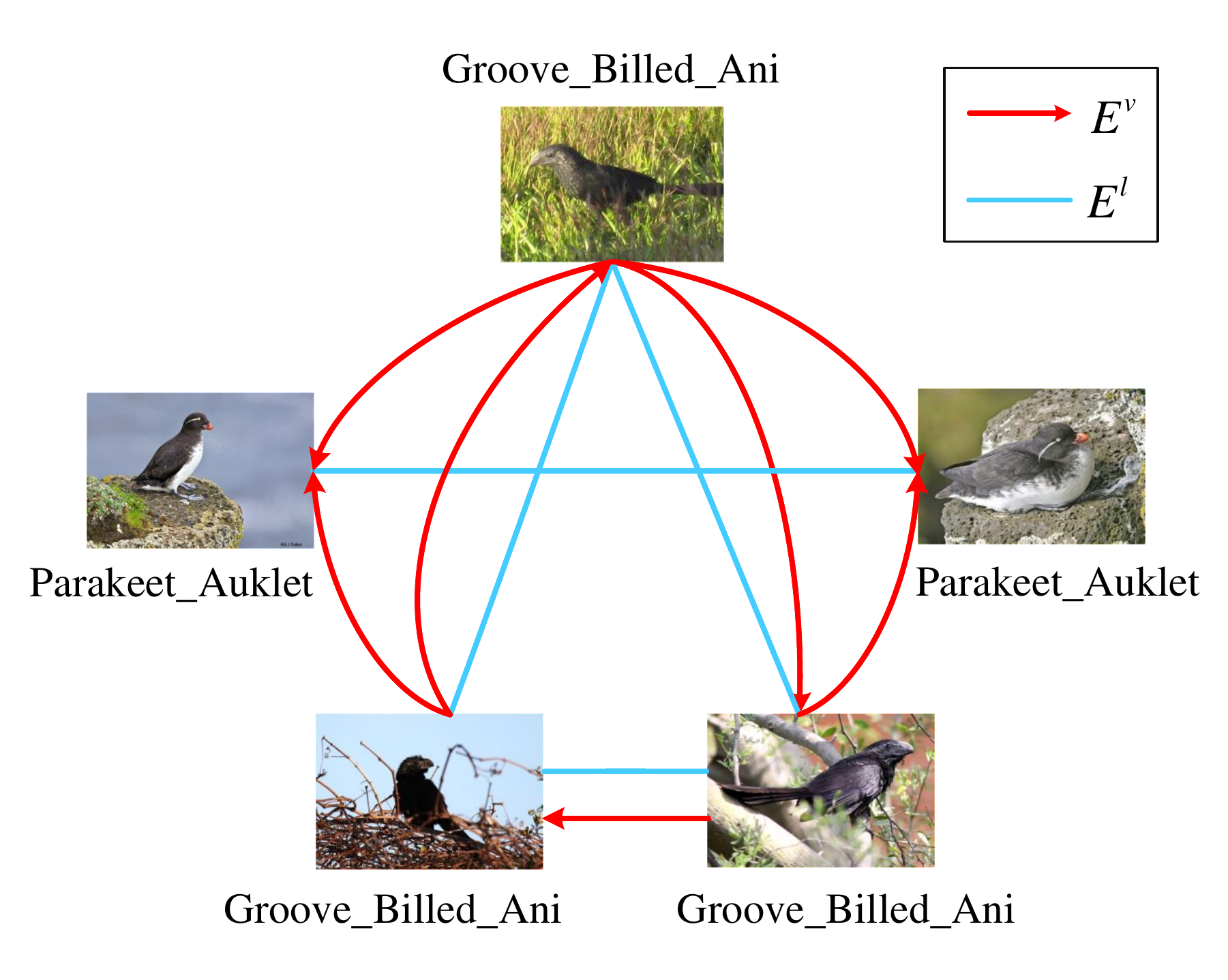}
\caption{ Illustration of Batch Graph, where $E^v$ denotes the visual-type edge set and $E^l$ represents the semantic-type edge set.
}
\label{fig:BG}
\end{figure}

To overcome these limitations,
in this paper, we rethink to explore sample relationships and %based Transformer  and
% In addition, the traditional Transformer assigns the correlation score to all the input images, which causes the learned image features to endure irrelevant image information.
% In conclusion, we argue that it is crucial to \emph{consider both visual and semantic correlation and eliminate the interference of irrelevant information} in feature representation learning.
propose a novel Batch-Graph Transformer (BGFormer) for robust data representation and learning.
The key aspects of the BGFormer are twofold.
\textit{First}, it employs a flexible graph model, termed Batch Graph to deeply model the relationships of samples within each mini-batch.
In particular, the proposed batch graph encodes {both} \textit{visual} and \textit{semantic/label} relationships of samples simultaneously in a unified model, as illustrated in Figure~\ref{fig:BG}.
Also, instead of utilizing the information of full relationships, our batch graph explores neighborhood relationships among samples by borrowing the idea of sparse graph representation~\cite{roweis2000lle,zhuang2012sparse,jiang2013glpca} which thus perform robustly, w.r.t., noisy samples.
\textit{Second}, based on the built batch graph, we then develop a novel graph learning mechanism, termed Batch-Graph Transformer (BGFormer), which aims to learn contextual representations for samples in a mini-batch by conducting message passing on the batch graph, as illustrated in Figure~\ref{fig:BGT}.
The core of BGFormer block is the dual Structure-constrained Self-attention (SSA) to simultaneously capture
visual and semantic relationships for data representation.

% Specifically, BGFormer mainly consists of batch visual graph attention, label graph attention and feature propagation module. It constructs visual and semantic/label correlation graph representing the different relations among all the input images in mini-batch separately. According to the multi-relational graph, then, we implement the feature propagation.
% In this process, the aggregation of irrelevant information may suppress the representation of the primary information, making the learned image feature representation under-focused.
% To this end, we introduce sparse graph representation approach to avoid this situation by aggregating solely the top-$K$ most relevant image neighbor node information.

To validate the effectiveness of the proposed BGFormer, we apply it to the metric learning tasks and develop a new deep metric learning approach.
As shown in Figure~\ref{fig:metric_learning_model}, our proposed metric learning framework is composed of backbone network, BGFormer module, hyperbolic embedding module and metric learning.
We first utilize the backbone network to extract the initial features for data samples.
Then, the proposed BGFormer is introduced to learn the more discriminative and richer representations by capturing the dependence of data samples.
Finally, we utilize the hyperbolic embedding module with shared parameters to map the learned sample features into hyperbolic space for metric learning tasks, as used in work~\cite{hou2022batchformer,ermolov2022Hyp}.

Overall, the main contributions of this work are summarized as follows:
\begin{itemize}
\item We rethink to deeply explore sample relationships
and build a flexible batch graph model to encode both visual and label interactions among different samples in each mini-batch.

\item
We design a novel BGFormer to explore sample relationships for robust and discriminative data representation in each mini-batch.
The proposed BGFormer is simple and provides a general plug-and-play module for data representation and learning tasks.

\item We integrate BGFormer into a deep metric learning scheme and propose an effective end-to-end metric learning approach.
Extensive experiments on four public popular datasets demonstrate the effectiveness and superiority of the proposed BGFormer.

\end{itemize}

\section{Related Works}
%In this section, we provide a brief review of visual Transformer, metric learning and graph representation learning methods.

\textbf{Vision Transformer.}
Since the Transformer architecture is proposed by Vaswani et al.~\cite{vaswani2017attention} in 2017, it has been widely used in the field of computer vision (CV) due to its excellent modeling ability, including image classification~\cite{liu2021swin,rao2021dynamicvit,wang2022few}, object re-identification~\cite{jiang2021ph,he2021transreid}, object detection~\cite{carion2020end}, object tracking~\cite{chen2021transformer,fu2022sparsett}, etc.
For example, Carion et al.~\cite{carion2020end} propose a Transformer encoder-decoder architecture, termed DEtection TRansformer (DETR for short), which drops multiple hand-designed components and builds the fully end-to-end object detector for object detection task.
Liu et al.~\cite{liu2021swin} design a hierarchical vision Transformer, which generates the feature representation with the shifted windows for image classification and other vision tasks.
He et al.~\cite{he2021transreid} introduce a pure Transformer-based object ReID framework, named TransReID, which splits each image as a sequence of patches and then feeds into TransReID to encode the feature representations for object ReID task.
%Wang et al.~\cite{wang2021transformer} design a Siamese-like tracking pipeline based on the transformer architecture, which use two parallel branches, i.e., encoder branch and decoder branch, to bridge search and template frame for object tracking task.
Zhang et al.~\cite{Zhang2022DGMN2} propose a Transformer-based backbone network based on dynamic graph message passing network for classification pretraining, object detection, instance and semantic segmentation.
These works adopt the Transformer as enhanced feature extractor and get better performance.

On the other hand, some metric learning~\cite{el2021IRT, ermolov2022Hyp} models are also developed based on the Transformer network. Specifically, El-Nouby et al.~\cite{el2021IRT} directly introduce the Transformer architecture to obtain the image feature representation for metric learning task. Ermolov et al.~\cite{ermolov2022Hyp} propose to combine the Transformer network and hyperbolic distance for obtaining the better results of metric learning.
%%%%
These works divide an image into a sequence of patches as the input of Transformer, but only model the contextual correlation of intra-image well. Seldom existing algorithms consider modeling relationships of inter-images using Transformer architecture.
In this work, we propose a novel image-level Transformer network, termed Batch-Graph Transformer (BGFormer) which captures the visual and sematic/label relationship between all the input images in each mini-batch by constructing batch graph.
Therefore, we can achieve better feature representation learning for high-performance classification.

\textbf{Metric Learning.}
With the help of neural networks, the deep metric learning~\cite{wu2017Margin, kim2018ABE, roth2019mic, wang2020XBM, teh2020PN} achieves significant improvements compared with traditional methods.
% that aims to employ various techniques or model architectures to obtain better data sample feature embeddings.
%In recent year, there are many methods to solve this problem, which can be briefly summarized into two categories: distance metric learning and deep metric learning.
%The distance metric learning~\cite{hadsell2006dimensionality,zhai2018NSmax,opitz2018BIER,ge2018HTL,qian2019ST,wang2019MS,kim2020PA} aims to design various loss functions to constrain the image sample of same class to be closer and different classes to be farther, including contrastive loss, normalized softmax loss, adversarial loss, hierarchical triplet loss, softTriple loss, proxy-anchor loss and so on.
%Unlike it, the goal of deep metric learning~\cite{wu2017Margin,kim2018ABE,roth2019mic,wang2020XBM,teh2020PN} is usually to obtain better data sample feature embeddings through various techniques or model architectures.
For example, Wu et al.~\cite{wu2017Margin} propose a distance weighted sampling strategy to uniformly extract samples according to the relative distance between samples for better embedding representation.
%i.e., distance weighted sampling, which uniformly extracts samples according to the relative distance between samples to reduce the variance of gradient and thus obtain a better embedding representation without considering the loss function.
Kim et al.~\cite{kim2018ABE} develop an attention-based ensemble based on multiple attention mask to improve the diverse of learners in feature embeddings for deep metric learning.
Roth et al.~\cite{roth2019mic} propose a model containing two encoder to explicitly learn the latent features shared between object classes in order to improve the performance of metric learning.
%, where the primary class encoder is trained with ground-truth labels and the auxiliary encoder is learned by a surrogate task.
Wang et al.~\cite{wang2020XBM} propose a cross-batch memory (XBM) mechanism to update the image features of recent mini-batches by memorizes the image embeddings of past mini-batches.
%%%%
Although good results can be obtained, however, these works adopt the CNN as their feature extractor which may obtain sub-optimal results only.
In this work, we demonstrate the effectiveness of Transformer architecture by conducting experimental analysis and extend it to Batch-Graph Transformer to explore the sample relationship among all the input images in each mini-batch for more robust and discriminative feature representations.

\textbf{Graph Representation Learning.}
The goal of graph representation learning~\cite{velickovic2017graph, zhuang2012sparse,roweis2000lle,ying2021graphormer,chen2022sat} is to project the nodes of graph to vector representations and retain as much topological information as possible.
In early years, local sparse and low-rank graph shallow learning models have been widely studied~\cite{roweis2000lle,zhuang2012sparse,jiang2013glpca}, such as,
Jiang et al.~\cite{jiang2013glpca} propose Graph Laplacian PCA by incorporate local $k$-NN graph learning into PCA. In recent years, graph neural networks (GNNs) and graph Transformers have  been studied.
For example, Kipf et al.~\cite{kipf2016semi} propose Graph Convolutional Network (GCN), which applies convolution operation to graphs and models the local structure and node features of learning graphs through approximate analysis in the graph spectral domain.
Veli{\v{c}}kovi{\'c} et al.~\cite{velickovic2017graph} develop graph attention network (GAT), which computes the hidden representations of each node in the graph and follows a self-attention strategy to aggregate its neighbors.
Jiang et al.~\cite{GECN} propose a Graph elastic Convolutional Network (GeCN) for local sparse graph data representation learning.
Ying et al.~\cite{ying2021graphormer} develop a Graphormer with centrality and spatial encodings to obtain the good results on graph representation learning.
Chen et al.~\cite{chen2022sat} propose a Structure-Aware Transformer to incorporate the structural information into the original self-attention.

In recent years, good performance has also been obtained by employing GCNs/GNNs on various computer vision tasks. %~\cite{qi2018learning, yan2019learning, chen2021learning, jiang2021ph, xie2021scale}.
For example, Qi et al.~\cite{qi2018learning} propose a Graph Parsing Neural Network (GPNN) based on the message passing neural network to provide a HIO representation for spatial and spatial-temporal domains.
Gao et al.~\cite{gct_cvpr19} design a Graph Convolutional Tracking (GCT) to model structured representation of historical target exemplars for visual tracking.
Chen et al.~\cite{chen2021learning} propose a multi-label recognition model based on Graph Convolutional Networks to learn inter-dependent class-level representation.
Xie et al.~\cite{xie2021scale} develop a scale-aware graph neural network (SAGNN) to learn the cross-scale relations between support-query image pairs for few-shot semantic segmentation.
%%%%
Differently, in this work, we introduce batch graph to model both visual and semantic/label relationships between all the input images in each mini-batch and develop a Batch-Graph Transformer for image-level representation and learning.

%BG figure

\section{Methodology}

\label{sec:sec3}
In this section, we propose a novel Batch-graph Transformer (BGFormer) model to deeply capture the relationships of samples in the mini-batch for data representation.
BGFormer has two aspects, i.e., i) batch graph construction and
 ii) representation learning on batch graph, as introduced below in detail.

%develop a novel Batch-Graph Transformer architecture, which is composed of a stack of multiple Batch-Graph Transformer block.
%The Batch-Graph Transformer block aims to mine the image sample relationships by considering the multiple relations between samples in each batch for the more robust and discriminative image feature representations.
%The more detailed description is provided below.

\subsection{Building Batch Graph}

To fully encode the relationships of samples in each mini-batch, we  build a dual-relational graph $G (V, E^v, E^l)$
which encodes
both visual and label correlations among different data samples, as illustructed in Figure~\ref{fig:BG}.
%After that it is fed into the Batch-Graph Transformer block for feature propagation.
% To be specific, the detailed nodes and edges of multi-relational graph are introduced as follow.

\textbf{Nodes. }
Each node $v_i$ in the batch graph $G (V, E^v, E^l)$ denotes each sample in the mini-batch. Let $f_i\in \mathbb{R}^{C}$ denotes the initial CNN feature descriptor  for sample node $v_i \in V$.
Let $F = \{f_1, f_2, \cdots, f_B \} \in \mathbb{R}^{B \times C}$ represents the collection of node's initial features where $B=|V|$ is the total number of samples (i.e., batch size) in a mini-batch size.
In our experiments, we empirically set the batch size to 900 and thus we have $|V| = 900$. % Thus, the size of nodes is 900.

\textbf{Edges.} The edge set in our batch graph involves
two types of edges, i.e., $E^v$ encodes the visual content relationships and $E^l$ represents the label relationships among samples.
For visual-type edges $E^v$, we adopt the neighboring connection strategy\footnote{Some more compact sparse graphs~\cite{zhuang2012sparse,GECN} can also be adopted here.}.
To be specific,  we first learn the affinities/similarities among data samples as
$
S = FF^T
$
%where $F\in \mathbb{R}^{B\times C}$
as suggested in the simplified self-attention~\cite{guo2022beyond}.
%are obtained by conducting two different linear projections on features $F$ respectively, as suggested in self-attention~\cite{}.
Then, each node $v_i\in V$  connects to the node $v_j\in V$ that are belong to its $k$ nearest neighborhood, i.e., $v_j \in \mathcal{N}_k(v_i)$, as shown in Figure~\ref{fig:BG}.
We can use the weight adjacency matrix $A^v\in \mathbb{R}^{B\times B}$ to represent visual-type edge connections which is defined as follows,
\begin{equation}
A^v_{ij}=\begin{cases}
  &s_{ij}, \text{ if } v_j \in \mathcal{N}_k(v_i)\\
  &0, \text{ otherwise}
\end{cases}
\label{equ:eq1}
\end{equation}
where $\mathcal{N}_k$ denotes the $k$ nearest neighborhoods of node $v_i$.

In addition to visual correlations, we also represent the label correlations of data samples via label-type edge set $E^l$. To be specific,
each edge $e^l_{ij}\in E^l$ connects node $v_i$ and $v_j$ if they have the identical class label.
We can use an adjacency matrix $A^l\in \mathbb{R}^{B\times B}$ to encode the label edge connections which is defined as
\begin{equation}
A^l_{ij}=\begin{cases}
  &1, \text{ if } l(v_i)=l(v_j) \\
  &0, \text{ otherwise.}
\end{cases}
\label{equ:eq2}
\end{equation}
where $l(v_i)$ denotes the label of sample node $v_i$.

\subsection{Batch-Graph Transformer}

\begin{figure}[!t]
\centering
\includegraphics[width=0.49\textwidth]{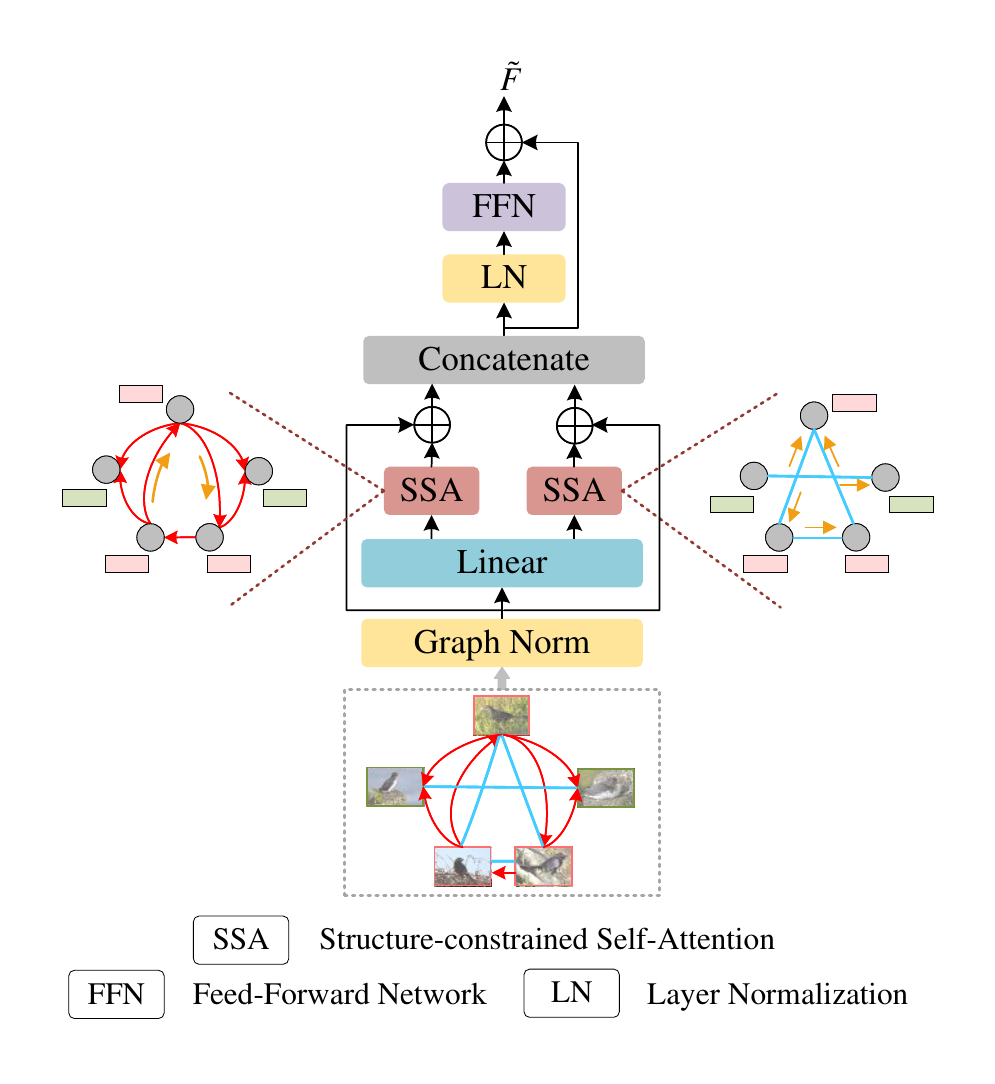}
\caption{ Illustration of the proposed BGFormer block.
}
\label{fig:BGT}
\end{figure}

% In this subsection, we introduce the detailed architecture of the proposed BGFormer.
Overall,  the proposed
BGFormer module involves  multiple BGFormer encoders and a layer-norm (LN) is used between different blocks, as shown in Figure~\ref{fig:BGT}.
Different from regular Transformer encoder, our BGFormer encoder first leverages two structure-constrained self-attention blocks to propagate information across neighboring samples (tokens) and  then fuses them together, followed by LN, FFN and residual operation to output the final data representations.

Specifically, based on the above batch graph representation $\{F, A^v, A^l\}$, we first conduct Structure-constrained Self-attention (SSA) on batch graph as
\begin{align}
% & \tilde{A}^v = \frac{A_{ij}^v-min(A_j^v)}{max(A_j^v)-min(A_j^v)} \\
& \tilde{F}^v = F + Norm (A^v) FW \\
 & \tilde{F}^l = F + Norm (A^l) FW
\label{equ:eq4}
\end{align}
where $Norm(\cdot)$ denotes the graph normalization operation, such as linear normalization, softmax and symmetric Laplacian normalization~\cite{kipf2016semi}.
$W$ denotes the weight of linear projection which is shared on two SSA blocks to achieve information communication across two blocks.
In this paper, for visual  affinity $A^v$, we use the linear normalization as follows,
\begin{align}
Norm(A^v) = \frac{A^v_{ij}-\min_{i,j} \{A^v_{ij}\}}{\max_{i,j}\{A^{v}_{ij}\} - \min_{i,j}\{A^{v}_{ij}\}}
\label{equ:eq5}
\end{align}
For label   affinity $A^l$, we use the symmetric Laplacian normalization as
\begin{align}
Norm(A^l) = D^{-\frac{1}{2}}(A^l+I)D^{-\frac{1}{2}}
\label{equ:eq6}
\end{align}
where $I$ denotes the identity matrix and $D$ is a diagonal matrix with $D_{ii} = \sum_{j}A^l_{ij}$.

After the above SSA module, we then fuse them together by simply combing them as
\begin{align}
 \hat{F} = \lambda \tilde{F}^v + (1-\lambda) \tilde{F}^l
 \label{equ:eq7}
\end{align}
where $\lambda \in \{0, 1\}$ denotes the balanced hyper-parameter.
Finally,  we introduce residual connection and feed-forward network (FFN) on $\hat{F}$ to output the final data representation as follows,
\begin{align}
\tilde{F} = \hat{F} + FFN(LN(\hat{F}))
\label{equ:eq8}
\end{align}
where $LN(\cdot)$ denote layer normalization and $FFN(\cdot)$ is composed of two fully-connected layers.

%BGT figure

% model figure
\begin{figure*}[!htbp]
\centering
\includegraphics[width=1\textwidth]{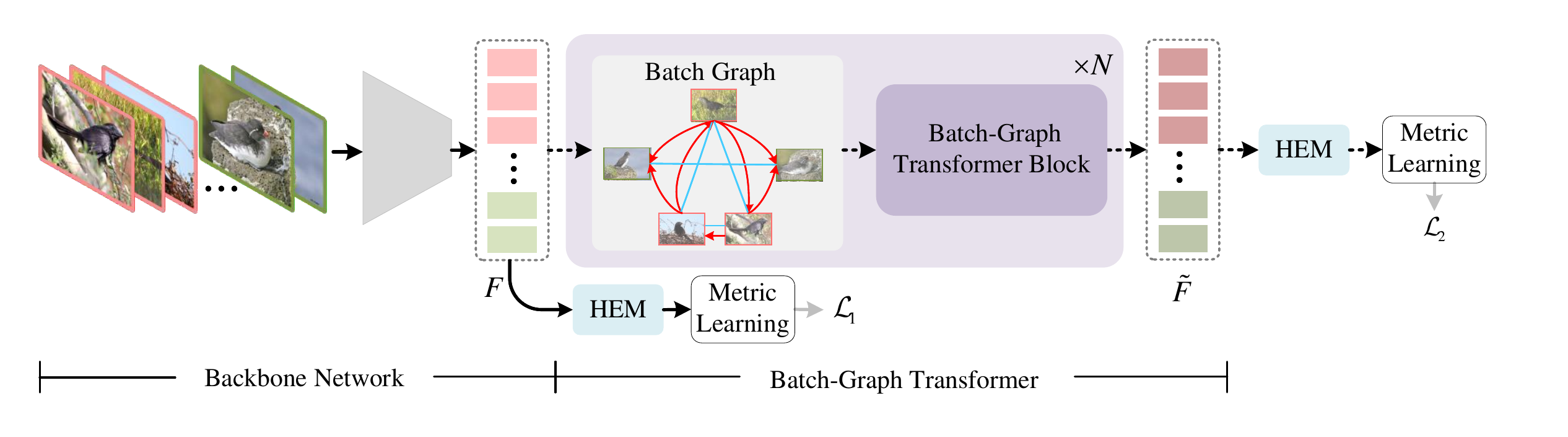}
\caption{Overview of the proposed Batch-Graph Transformer (BGFormer) and application in metric learning tasks.
Specifically, we first use a backbone network to generate the initial feature embeddings for samples in each mini-batch.
After that, we adopt the proposed BGFormer module to fully explore the cues of sample relationships (visual, label) to enrich
sample data representations.
Finally, we adopt a hyperbolic embedding module (HEM) to map both initial and enhanced sample representations $F, \widetilde{F}$ into the hyperbolic space for metric learning tasks, as discussed in \S 4.
The black dotted line indicates it is not performed in the test phase.
}
\label{fig:metric_learning_model}
\end{figure*}

\section{Application: Metric Learning}
To verify the effectiveness of our proposed BGFormer, we apply it on the metric learning tasks.
As shown in Figure~\ref{fig:metric_learning_model}, the BGFormer-based metric learning framework mainly consists of backbone network, BGFormer module, hyperbolic embedding module and metric learning.

Let $X = \{(x_i, y_i)\}_{i=1}^B$ be the inputs of backbone network, where $y_i$ is the corresponding label of image $x_i$ and $B$ denotes the number (i.e., batch size) of images in each mini-batch.
The backbone network is a feature extraction network, which can be CNNs~\cite{he2016Resnet,ioffe2015InceptionBN} (e.g., ResNet-50) or Transformer~\cite{dosovitskiy2020ViT,touvron2021DeiT,caron2021DINO} (e.g., ViT, DeiT and DINO).
In this work, we  employ ViT-S~\cite{dosovitskiy2020ViT} with adding the global average pooling (GAP) layer as the backbone network to generate the initial image feature descriptors $F \in \mathbb{R}^{B \times C}$.
Its parameters are initialized by using the pre-trained model on ImageNet-21K~\cite{deng2009imagenet} for classification tasks.
Then, we feed the initial image features $F$ into the proposed BGFormer module to fully explore the relationships of images in each mini-batch for learning the enhanced representations $\tilde{F} \in \mathbb{R}^{B \times C}$.
After that, based on the image features $F$ and $\tilde{F}$, we use the Hyperbolic Embedding module (HEM) involving fully-connected (FC) layer and hyperbolic mapping, which embeds them into the hyperbolic space to calculate the hyperbolic distances between pairs of samples~\cite{ermolov2022Hyp}.
Finally, we leverage the pairwise cross-entropy loss function which defined on pair-wise distance~\cite{ermolov2022Hyp} to achieve the metric learning task.

Since the BGFormer module is not utilized in the testing stage, to transfer the knowledge of training process to the test stage, we derive two Hyperbolic Embedding (HEM) and metric learning branches
for the outputs $F$ and $\widetilde{F}$ respectively, as suggested in work~\cite{hou2022batchformer}. Both learning branches share  the same parameters
to achieve the knowledge communication.
Let $\mathcal{L}_1$ and $\mathcal{L}_2$
denote the pairwise cross-entropy loss for these two branches.
Then, the total loss of our method is formulated as
% The pairwise cross-entropy loss can be denoted as $\mathbb{L}_1$ and $\mathbb{L}_2$ on $F$ and $\tilde{F}$, respectively.
% Finally, we can obtain the total loss function, which can be written as
%
\begin{equation}
\mathcal{L}_{total} = \alpha \mathcal{L}_1 + (1-\alpha) \mathcal{L}_2
\label{equ:eq12}
\end{equation}
where $\alpha \in \{0, 1\}$ is a balanced hyper-parameter.

\begin{table*}[!htbp]
\centering
\setlength\tabcolsep{3.2pt}
\begin{tabular}{|l|l|llll|llll|llll|llll|}
\hline
\multirow{2}{*}{Method} & \multirow{2}{*}{dim} & \multicolumn{4}{c|}{CUB-200-2011 (K)} & \multicolumn{4}{c|}{Cars-196 (K)} & \multicolumn{4}{c|}{SOP (K)} & \multicolumn{4}{c|}{In-Shop (K)}  \\%\cline{3-18}
            &      &1    &2    &4    &8     &1    &2    &4    &8    &1    &10   &100  &1000  &1    &10   &20   &30  \\ \hline
SM~\cite{suh2019SM}           &512 &56.0 &68.3 &78.2 &86.3  &83.4 &89.9 &93.9 &96.5 &75.3 &87.5 &93.7 &97.4  &90.7 &97.8 &98.5 &98.8 \\
HTL~\cite{ge2018HTL}          &512 &57.1 &68.8 &78.7 &86.5  &81.4 &88.0 &92.7 &95.7 &74.8 &88.3 &94.8 &98.4  &80.9 &94.3 &95.8 &97.2 \\
A-BIER~\cite{opitz2018BIER}   &512 &57.5 &68.7 &78.3 &86.2  &82.0 &89.0 &93.2 &96.1 &74.2 &86.9 &94.0 &97.8  &83.1 &95.1 &96.9 &97.5 \\
ABE~\cite{kim2018ABE}         &512 &60.6 &71.5 &79.8 &87.4  &85.2 &90.5 &94.0 &96.1 &76.3 &88.4 &94.8 &98.2  &87.3 &96.7 &97.9 &98.2 \\
NSoftmax~\cite{zhai2018NSmax} &512 &61.3 &73.9 &83.5 &90.0  &84.2 &90.4 &94.4 &\underline{96.9} &78.2 &90.6 &96.2 & -    &86.6 &97.5 &98.4 &98.8 \\
SoftTriple~\cite{qian2019ST}  &512 &65.4 &76.4 &84.5 &90.4  &84.5 &90.7 &94.5 &\underline{96.9} &78.6 &86.6 &91.8 &95.4  & -   & -   & -   & -   \\
MS~\cite{wang2019MS}          &512 &65.7 &77.0 &86.3 &91.2  &84.1 &90.4 &94.0 &96.5 &78.2 &90.5 &96.0 &98.7  &89.7 &97.9 &98.5 &98.8 \\
XBM~\cite{wang2020XBM}        &512 &65.8 &75.9 &84.0 &89.9  &82.0 &88.7 &93.1 &96.1 &79.5 &90.8 &96.1 &98.7  &89.9 &97.6 &98.4 &98.6 \\
%HORDE~\cite{jacob2019HORDE}   &512 &66.8 &77.4 &85.1 &91.0  &86.2 &\textbf{91.9} &\underline{95.1} &\underline{97.2} &80.1 &91.3 &96.2 &98.7  &90.4 &97.8 &98.4 &98.7 \\
Proxy-Anchor~\cite{kim2020PA} &512 &68.4 &79.2 &86.8 &91.6  &\textbf{86.1} &\underline{91.7} &\underline{95.0} &\textbf{97.3} &79.1 &90.8 &96.2 &98.7  &91.5 &98.1 &98.8 &\textbf{99.1} \\
%ProxyNCA++~\cite{teh2020PN}   &512 &69.0 &79.8 &87.3 &92.7  &86.5 &\textbf{92.5} &\textbf{95.7} &\textbf{97.7} &80.7 &92.0 &96.7 &98.9  &90.4 &98.1 &98.8 &\underline{99.0} \\
IRT$_R$~\cite{el2021IRT}      &384 &76.6 &85.0 &91.1 &94.3  & -   & -   & -   & -   &84.2 &93.7 &97.3 &99.1  &91.9 &98.1 &98.7 &\underline{98.9} \\
Margin~\cite{wu2017Margin}    &128 &63.9 &75.3 &84.4 &90.6  &79.6 &86.5 &91.9 &95.1 &72.7 &86.2 &93.8 &98.0  & -   & -   & -   & -   \\
FastAP~\cite{cakir2019FastAP} &128 & -   & -   & -   & -    & -   & -   & -   & -   &73.8 &88.0 &94.9 &98.3  & -   & -   & -   & -   \\
MIC~\cite{roth2019mic}        &128 &66.1 &76.8 &85.6 & -    &82.6 &89.1 &93.2 & -   &77.2 &89.4 &94.6 & -    &88.2 &97.0 & -   &98.0 \\
Hyp-ViT~\cite{ermolov2022Hyp}$^*$ &128 &\underline{84.0} &\underline{90.2} &\underline{94.2} &\underline{96.4}  &82.7 &89.7 &93.9 &96.2 &\underline{85.5} &\underline{94.9} &\underline{98.1} &\underline{99.4}  &\underline{92.7} &\underline{98.4} &\underline{98.9} &\textbf{99.1} \\ \hline
BGFormer$^*$                      &128 &\textbf{86.6} &\textbf{92.0} &\textbf{94.9} &\textbf{96.7} &\underline{86.0} &\textbf{91.9} &\textbf{95.2} &\textbf{97.3} &\textbf{86.4} &\textbf{95.4} &\textbf{98.4} &\textbf{99.5} &\textbf{92.8} &\textbf{98.5} &\textbf{99.0} &\textbf{99.1}             \\\hline
%            &      &      &      &      &                  &      &      &      &              &      &      &      &         &      &      &      &              \\\hline
\end{tabular}
\caption{Recall@K metric for four datasets, where 'Dim' column represents the dimensionality of the final embeddings of model.
The different backbone network is used by these methods, such as ResNet-50~\cite{he2016Resnet} (NSoftmax, Margin, FastAP, MIC), GoogleNet~\cite{szegedy2015GoogleNet} (SM, A-BIER, ABE), Inception~\cite{ioffe2015InceptionBN} (HTL, SoftTriple, MS, XBM, Proxy-Anchor) and Transformer~\cite{dosovitskiy2020ViT,touvron2021DeiT} (IRT$_R$, Hyp-ViT).
$^*$ means it is pretrained on the larger ImageNet-21k~\cite{deng2009imagenet} dataset.
The \textbf{bold} and \underline{underline} denote the best two results.
}
\label{tab:tb1}
\end{table*}

\section{Experiments}
%In this section, we conduct extensive experiments on four public popular benchmarks for category-level retrieval to validate the effectiveness of the proposed method.
%Below, we first show the detailed settings of four datasets, model implementation and training details in turn.
%Next, we follow the widely adopted evaluation protocol and compare the performances of recent state-of-the-art methods on four datasets.
%Finally, we conduct ablation studies on the proposed method, including parameter analysis, component analysis, etc.
%These are described in more detail in the following sub-sections.

\subsection{Datasets and Evaluation Metrics}
\textbf{Datasets.}
We evaluate the proposed BGFormer based metric learning method on four public datasets, including \textbf{CUB-200-2011 (CUB)}~\cite{welinder2010caltech} (200 classes, 11788 images), \textbf{Cars-196 (Cars)}~\cite{krause20133d} (196 classes, 16185 images), \textbf{Stanford Online Product (SOP)}~\cite{oh2016deep} (22634 classes, 120053 images) and \textbf{In-shop Clothes Retrieval (In-Shop)}~\cite{liu2016deepfashion} (7986 classes, 52712 images), which are all splited into training set and testing set.
Concretely, by following works~\cite{el2021IRT, ermolov2022Hyp}, we select 100 classes (5864 images), 98 classes (8054 images) and 11318 classes (59551 images) from CUB, Cars and SOP respectively for training and the rest classes are used for testing.
For the In-Shop dataset, we select the first 3997 classes (25882 images) for training and split the remaining 3985 classes into a query set (14218 images) and a gallery set (12612 images) for testing.

\textbf{Evaluation Metrics.}
In this work, we utilize the widely adopted Recall@K evaluation metric to compare the proposed method with previous metric learning methods.
The Recall@K represents the ratio of the number of positive samples retrieved in the top-K results to the number of all positive samples in the testing set.
For a fair comparison, following these works~\cite{zhai2018NSmax, cakir2019FastAP, wang2020XBM, el2021IRT, ermolov2022Hyp},
for CUB and Cars, we set $K$ to 1, 2, 4 and 8.
For SOP, we set $K$ to 1, 10, 100 and 1000.
For In-Shop, we set $K$ to 1, 10, 20 and 30.

% tsne
\begin{figure*}[!htb]
\centering
\includegraphics[width=0.8\textwidth]{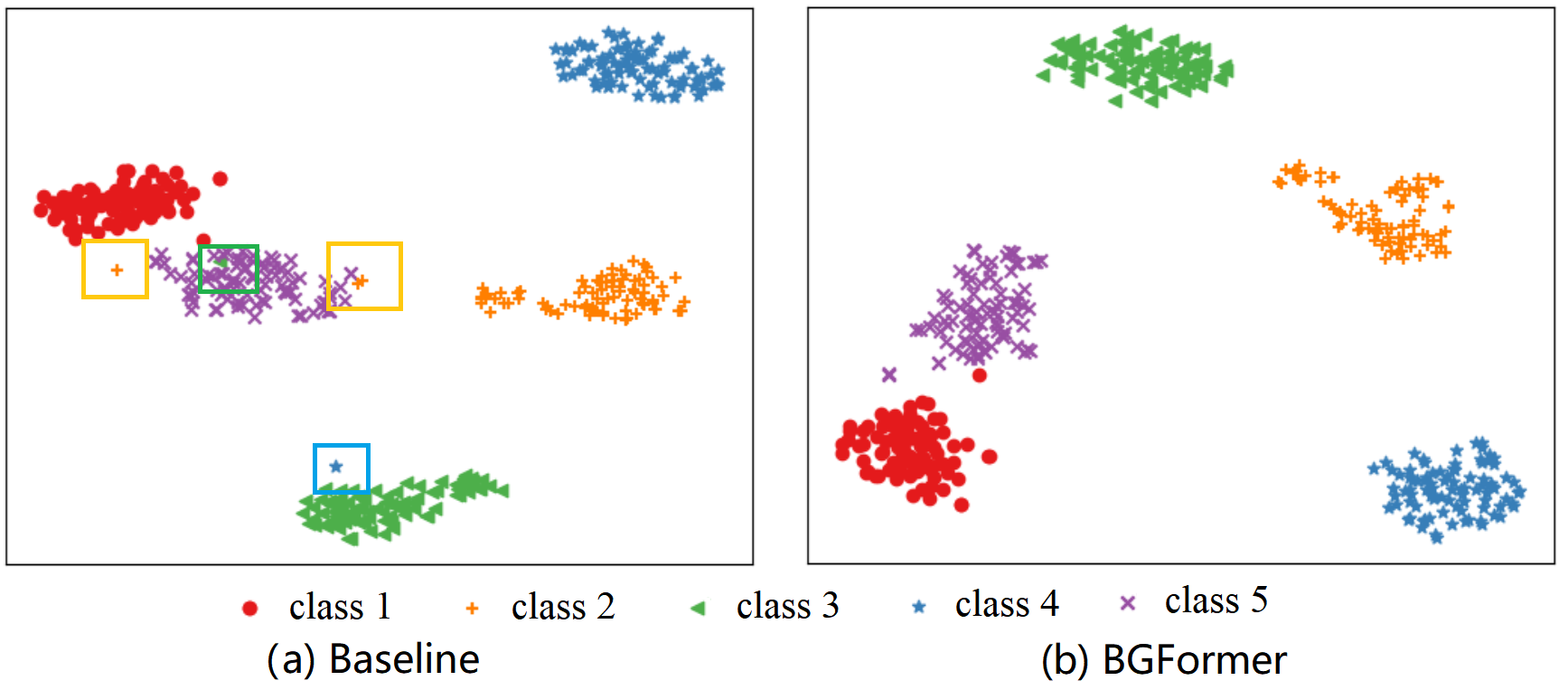}
\caption{2D t-SNE~\cite{van2008visualizing} visualization of feature representations learned by Baseline and BGFormer on Cars~\cite{krause20133d} dataset, respectively.
The boxes are the outliers.
Different colors indicate different classes.
}
\label{fig:tsne}
\end{figure*}

% rank figure
\begin{figure}[!htbp]
\centering
\includegraphics[width=0.48\textwidth]{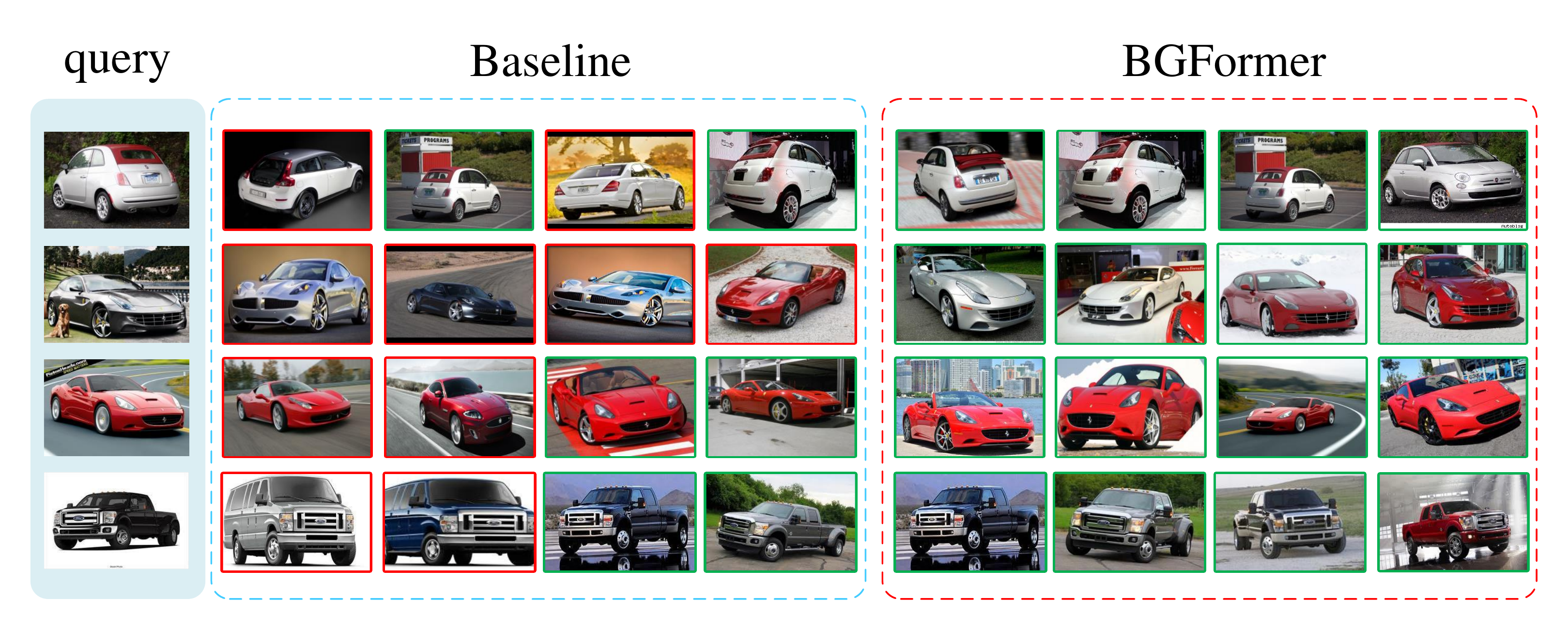}
\caption{Representive rank-4 results provided by the Baseline~\cite{ermolov2022Hyp} and our BGFormer on Cars~\cite{krause20133d} dataset, respectively. The \textcolor{green}{Green} boundary denotes true positive samples and \textcolor{red}{Red} is false positive samples.
}
\label{fig:qualitative_results}
\end{figure}

\subsection{Implementation Details}
In this work, we employ ViT-S~\cite{dosovitskiy2020ViT} network pre-trained on the large-scale ImageNet-21k~\cite{deng2009imagenet} as the backbone network for extracting the initial image feature representations.
%Then we use a classifier after the proposed Batch-Graph Transformer (BGFormer) module for metric learning task.
%Then we add an auxiliary hyperbolic embedding module (HEM) with with shared parameters before BGFormer module to enable the sample relationship information learned from the BGFormer module to be transfer to the backbone network and auxiliary HEM.
%In addition, we add an auxiliary classifier with shared parameters before BGFormer module to enable the sample relationship information learned from the BGFormer module to be transfer to the backbone network and auxiliary classifier.
We execute our proposed model with Python and train the whole network in an end-to-end way.

The resolution of all images in the CUB dataset is set as $256 \times 256$. Following the work~\cite{ermolov2022Hyp}, other datasets are set as $224 \times 224$. In the training phase, we randomly select 100 classes and 9 samples per class in each mini-batch, i.e., batch size $B = 900$. Different learning rates are set for the backbone network and proposed BGFormer module, i.e., 3e-5 and 7.5e-4 respectively. The AdamW~\cite{loshchilov2017decoupled} is adopted as optimizer, with the weight decay 5e-5 to optimize the whole network. Our model can be trained well with a total number of 400 epochs, and we test the checkpoint every 5 epochs. The checkpoint with the best results is saved as the final model. % It is worth noting that both the BGFormer module and the final classifier can be removed in the inference phase, with the help of an auxiliary classifier with shared parameters. That is to say, only the baseline network is used for testing.

\subsection{Comparison with State-of-the-Art Methods}
The comparison results on CUB-200-2011 (CUB)~\cite{welinder2010caltech}, Cars-196 (Cars)~\cite{krause20133d}, Stanford Online Product (SOP)~\cite{oh2016deep} and In-shop Clothes Retrieval (In-Shop)~\cite{liu2016deepfashion} datasets are summarized in Table~\ref{tab:tb1}.
We can clearly see that our proposed model outperforms all current state-of-the-art methods including CNN-based methods~\cite{wu2017Margin,ge2018HTL,opitz2018BIER,kim2018ABE,zhai2018NSmax,suh2019SM,qian2019ST,wang2019MS,cakir2019FastAP,roth2019mic,wang2020XBM,kim2020PA} and Transformer methods~\cite{el2021IRT,ermolov2022Hyp}. Note that our proposed approach achieves the best scores on all four settings on the CUB dataset. Specifically, our model improves the Hyp-ViT~\cite{ermolov2022Hyp} based on Transformer architecture and hyperbolic distance by +2.6\% in Recall@1 and +1.8\% in Recall@2. The IRT$_R$~\cite{el2021IRT} is also developed based on Transformer architecture, however, we exceed this approach by a large margin, i.e., +10.0\% in Recall@1 and +7.0\% in Recall@2. Compared with the Proxy-Anchor~\cite{kim2020PA} with the SOTA performance based on CNN architecture, we still obtain +18.2\% and +12.8\% improvements in Recall@1 and Recall@2 respectively.

From the experimental results on the Cars dataset, we can find that our model achieves the best performance on most of the evaluation metrics and the second best results on Recall@1. It beats the recent strong Hyp-ViT~\cite{ermolov2022Hyp} which is developed based on Transformer architecture and hyperbolic space, by +3.3\% in Recall@1 and +2.2\% in Recall@2. Similar findings and conclusions can be found in the SOP and InShop datasets. In addition, it is worth noting that our proposed model with more minor or the same low-dimension embeddings achieves the SOTA performance for all datasets.

Overall, our proposed model obtains the best performance on both more challenging fine-grained datasets like Cars and CUB, and large-scale datasets with enormous classes like SOP and InShop. These experimental results fully demonstrate the effectiveness of the proposed model on learning more discriminative feature representations.

\begin{table}[!tbp]
\centering
\setlength\tabcolsep{8pt}
\begin{tabular}{l|llll}
\hline
\multirow{2}{*}{Parameter} &\multicolumn{4}{c}{Recall @ K} \\ \cline{2-5} %\multicolumn{4}{c|}{CUB-200-2011 (K)} & \multicolumn{4}{c|}{Cars-196 (K)}   \\ \cline{3-18}
     &1    &2    &4    &8  \\ \hline   %&1    &2    &4    &8     \\ \hline
Default    &86.6 &92.0 &94.9 &96.7 \\ \hline  %&86.0 &91.9 &95.2 &97.3            \\ \hline
d=384      &86.6 &91.9 &94.8 &96.5 \\ %&84.7 &91.2 &94.9 &97.2  \\
%d=256      &85.9 &91.7 &94.5 &96.2   \\
d=64       &85.5 &91.4 &94.2 &96.4  \\
d=16       &85.0 &90.9 &94.5 &96.5   \\\hline
B=1200     &86.4 &91.8 &94.8 &96.7  \\ %&85.4 &91.6 &95.3 &97.5  \\
B=800      &86.3 &91.6 &94.5 &96.6  \\ %&85.5 &91.7 &95.3 &97.5  \\
B=400      &85.7 &91.4 &94.5 &96.2  \\ %&84.8 &90.7 &94.6 &97.1  \\
B=200      &85.6 &91.6 &94.4 &96.2   \\ \hline
\end{tabular}
\caption{Comparison results of different backbone networks on CUB~\cite{welinder2010caltech} dataset.
The default configuration of the proposed model is the dimension of final output $d=128$, batch size $B=900$.
}
\label{tab:tb2}
\end{table}

\textbf{Qualitative Visualization.}
%We show the top-4 ranking list of metric learning results for some query images on Cars~\cite{krause20133d} dataset, which can be found in supplementary material due to the limited space.
%We can clearly see that more true positives can be found using the proposed BGFormer.
As shown in Figure~\ref{fig:qualitative_results}, we give more visualizations of the top-4 ranking results on the Cars~\cite{krause20133d} dataset. The image with green boundary represents true positive samples and the red boundary denotes false positive samples. We can find that:
1). more true positives can be found in the proposed BGFormer.
2). the proposed BGFormer can accurately identify the same type of car with different colors, such as the 2-th and 4-th rows, which fully demonstrated the ability of BGFormer to close the intra-class gap.
3). the proposed BGFormer can well distinguish samples between different classes with little change, such as two types of cars in 2-th and 3-th rows.
In summary, we can obtain higher accuracy using the proposed BGFormer for the metric learning task.

\textbf{Representation Visualization.}
As shown in Figure~\ref{fig:tsne}, we show the 2D visualizations of the final image representations learned by the Baseline model~\cite{ermolov2022Hyp} and the proposed BGFormer on Cars~\cite{krause20133d} dataset, respectively. Different colors indicate different classes. Intuitively, we can find that the Baseline model can learn the discriminative image feature representation. It is because of the utilization of powerful ViT-S~\cite{dosovitskiy2020ViT} as the backbone network which is pre-trained on the large-scale ImageNet-21k dataset. However, the `outliers' are common to see from Figure~\ref{fig:tsne} (a) due to the baseline model only models the contextual correlation of intra-image but ignores the importance of the correlation of inter-images.
%%%%
In contrast, our proposed BGFormer significantly alleviates the issues of `outliers' and obtains more discriminative feature representations (the distribution of images is also clearer than our Baseline). This visualization further proves the effectiveness of the proposed BGFormer for learning more discriminative and robust image representations.

%We add the 2D visualizations of the final image representations learned by the Baseline model and the proposed BGFormer on Cars~\cite{krause20133d} dataset, respectively.
%The visualization can be found in supplementary material.
%We can observe that the feature representations learned by the proposed BGFormer have fewer outliers and the clearer distribution of images than our Baseline.
%Overall, this visualization further proves the effectiveness of the proposed BGFormer for learning more discriminative image representations.

% \usepackage{multirow}
\begin{table*}[!htb]
\centering
\setlength\tabcolsep{3.2pt}
\begin{tabular}{l|l|llll|llll|llll|llll}
\hline
\multirow{2}{*}{Methods} &\multirow{2}{*}{Backbone} & \multicolumn{4}{c|}{CUB-200-2011 (K)} & \multicolumn{4}{c|}{Cars-196 (K)} & \multicolumn{4}{c|}{SOP (K)} & \multicolumn{4}{c}{In-Shop (K)}  \\%\cline{3-18}
         &   &1    &2    &4    &8     &1    &2    &4    &8    &1    &10   &100  &1000  &1    &10   &20   &30  \\ \hline
Baseline  &ViT-S$^*$         &84.7 &90.9 &94.2 &96.5  &83.9 &90.2 &94.3 &96.9  &83.1 &93.5 &97.6 &99.3 &90.8 &97.9 &98.6 &98.8  \\ \hline
\multirow{4}{*}{BGFormer}&ResNet50 &45.2 &58.8 &71.6 &81.6  &45.6 &57.7 &69.8 &80.0  &73.1 &86.9 &94.4 &98.2 &76.0 &92.9 &95.0 &96.0\\
                    &DeiT-S        &78.2 &86.6 &91.8 &95.2  &85.3 &91.6 &95.2 &\textbf{97.5} &84.2 &94.0 &97.8 &99.3  &91.4 &98.1 &98.7 &98.9 \\
                    &DINO          &80.1 &87.9 &92.8 &95.4  &\textbf{87.7} &\textbf{93.1} &\textbf{96.0} &\textbf{97.5} &84.5 &94.2 &97.8 &99.3 &91.7 &98.1 &98.8 &99.0 \\
                    &ViT-S$^*$     &\textbf{86.6} &\textbf{92.0} &\textbf{94.9} &\textbf{96.7}  &86.0 &91.9 &95.2 &97.3 &\textbf{86.4} &\textbf{95.4} &\textbf{98.4} &\textbf{99.5} &\textbf{92.8} &\textbf{98.5} &\textbf{99.0} &\textbf{99.1}             \\\hline
\end{tabular}
\caption{Comparison results of different backbone networks on four datasets of metric learning task.
$^*$ means it is pretrained on the larger ImageNet-21k~\cite{deng2009imagenet} dataset and other methods are pre-trained on ImageNet~\cite{russakovsky2015imagenet} dataset.
In practice, we employ ViT-S as our backbone network.
The \textbf{bold} denotes the best results.
}
\label{tab:tb3}
\end{table*}

% \usepackage{multirow}
\begin{table*}[]
\centering
\setlength\tabcolsep{2.5pt}
\begin{tabular}{c|l|cccc|cccc|cccc|cccc}
\hline
\multirow{2}{*}{Model} &\multirow{2}{*}{Method} &\multicolumn{4}{c|}{CUB-200-2011(K)}  &\multicolumn{4}{c|}{Cars-196(K)}                                                         &\multicolumn{4}{c|}{SOP(K)}   &\multicolumn{4}{c}{In-Shop(K)}         \\ \cline{3-18}
  &                       &1    &2    &4    &8    &1    &2    &4    &8    &1    &10   &100  &1000 &1    &10   &20   &30   \\ \hline
1 &Baseline               &84.7 &90.9 &94.2 &96.5 &83.9 &90.2 &94.3 &96.9 &83.1 &93.5 &97.6 &99.3 &90.8 &97.9 &98.6 &98.8 \\ %\hline
2 &\ \ +BatchFormer       &85.2 &91.1 &94.5 &96.6 &84.1 &90.4 &94.5 &96.9 &83.8 &93.9 &97.9 &99.4 &91.6 &98.2 &98.8 &99.0 \\ %\hline
3 &\ \ +BGFormer(w/o LCG) &86.1 &\textbf{92.2} &\textbf{94.9} &96.6 &85.1 &91.1 &94.8 &97.2 &85.3 &94.7 &98.1 &99.4 &92.2 &98.4 &\textbf{99.0} &\textbf{99.1} \\ %\hline
4 &\ \ +BGFormer          &\textbf{86.6} &92.0 &\textbf{94.9} &\textbf{96.7} &\textbf{86.0} &\textbf{91.9} &\textbf{95.2} &\textbf{97.3} &\textbf{86.4} &\textbf{95.4} &\textbf{98.4} &\textbf{99.5} &\textbf{92.8} &\textbf{98.5} &\textbf{99.0} &\textbf{99.1} \\ \hline
\end{tabular}
\caption{Analysis experiments of different components of the proposed model, where `w/o LCG' represents the semantic/label correlation graph is removed.
The \textbf{bold} denotes the best results.
}
\label{tab:tb4}
\end{table*}

\subsection{Ablation Study}
In this section, we conduct extensive ablation studies to further investigate the influence of different modules or sets, such as embedding size, batch size, neighbor number, different backbone networks, etc.

\textbf{Analysis of Embedding Size.}
As we all know, the dimension of feature embedding affects the running speed and the overall accuracy. The larger scale of output dimension may carry more information and brings us better performance, but also leads to inefficiency issue and require more computation cost. In this part, we test different embedding dimensions to investigate their influence on our final performance, including $d=384, d=128, d=64$, and $d= 16$. As shown in Table~\ref{tab:tb2}, we can find that the performance decreases significantly when $d<128$ and is slightly improved when $d\ge 128$. Therefore, the embedding dimension $d=128$ is chosen as our default value in other experiments in this work.

\textbf{Effect of Batch Size.}
Different from the standard Transformer~\cite{dosovitskiy2020ViT, touvron2021DeiT, caron2021DINO} which extracts the patch-level features, in this work, we propose the BGFormer (Batch-Graph Transformer) to learn the image-level visual representations.
 %by considering the visual and label correlations in a graph.
Therefore, the batch size may be a key factor for the final performance.
As shown in Table~\ref{tab:tb2}, we analyze this parameter on the CUB~\cite{welinder2010caltech} dataset. It is easy to find that the overall performance benefits from the larger batch size $B$, and will be relatively stable when the batch size $B\ge 800$. When the batch size $B < 400$, the performance will drop slightly. In our experiments, we set $B=900$ to achieve a good balance between the performance and the hardware limitation.

\begin{figure}[!t]
\centering
\includegraphics[width=0.45\textwidth]{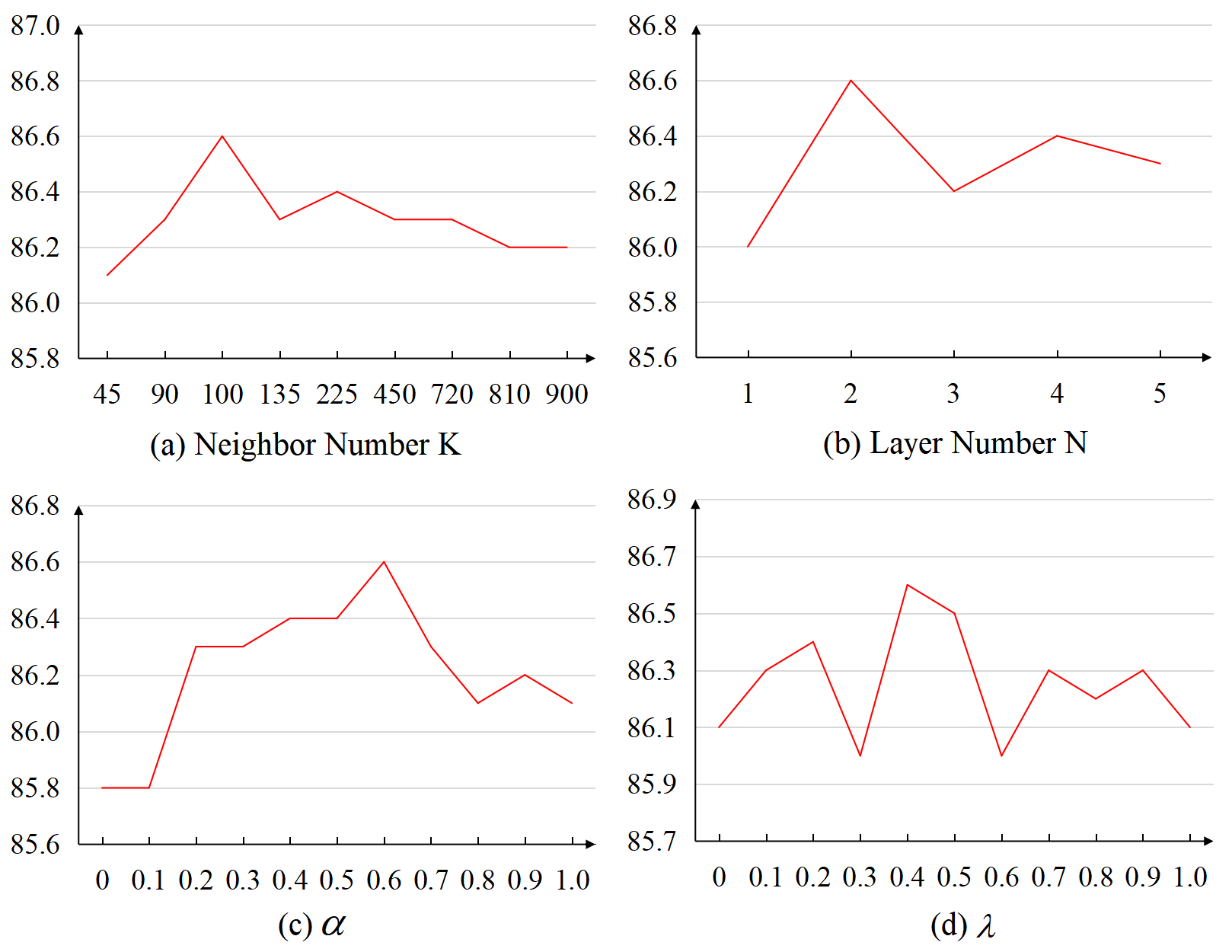}
\caption{Recall@1 metric for different parameter settings in the proposed model on CUB~\cite{welinder2010caltech} dataset.
%, including neighbor number $K$, the number $N$ of BGFormer layers and two balanced hyper-parameters $\alpha$, $\lambda$.
%We default set $K=100$, $N=2$, $\alpha=0.6$ and $\lambda=0.4$ in our experiment.
}
\label{fig:parm_Analysis}
\end{figure}

\textbf{Effect of Neighbor Number.}
The neighbor number $K$ controls the computational complexity of structure-constrained self-attention based on visual graph in the proposed BGFormer module.
The sparse operation can reduce the computational complexity of feature aggregation from $\mathcal{O}(B^2C)$ to $\mathcal{O}(BKC)$, where $K$ is a constant and $K < B$.
To this end, we study the influence of different number of neighbors, as shown in Figure~\ref{fig:parm_Analysis} (a).
We can see that the model performance increases with the decrease of neighbor number until $K=100$.
We speculate that the fewer irrelevant image samples are considered in feature aggregation, which reduces the impact of negative samples and makes the image features learned from the proposed model more discriminative.
Therefore, we default set $K=100$ in all our experiments.

\textbf{Analysis of Backbone Network.}
As shown in Table~\ref{tab:tb3}, we report the performance of four public datasets that extract initial image features based on different backbones, including ResNet50~\cite{he2016Resnet, ioffe2015InceptionBN}, DeiT-S~\cite{touvron2021DeiT}, DINO~\cite{caron2021DINO}, and ViT-S~\cite{dosovitskiy2020ViT}.
We can find that
1). the performance of Transformer architecture is obviously better than the ResNet50 network.
2). when equipping our proposed BGFormer with different backbones, i.e., DeiT-S and DINO, and ViT-S, most of our results are also better than the baseline method.
3). Our proposed BGFormer achieves the best performance when ViT-S is employed as the backbone network, except for the Cars~\cite{krause20133d} dataset.
However, it achieves sub-optimal performance on the Cars dataset.
As a result, ViT-S is adopted as the default backbone of our model in all our experiments.

\textbf{Analysis of Different Components.}
To verify the effectiveness of each module in our proposed model, we conduct extensive ablation analysis as shown in Table~\ref{tab:tb4}. To be specific, the baseline (Model 1) adopts the standard ViT-S~\cite{dosovitskiy2020ViT} model for deep metric learning as suggested in~\cite{ermolov2022Hyp}. Based on this model, we directly add BatchFormer (Model 2) to model the relationships among samples in each mini-batch by following~\cite{hou2022batchformer}. Instead of using full relations, we develop BGFormer (w/o LCG) by adding sparse graph representation to avoid the impact of negative samples (Model 3). In addition, we add structure-constrained self-attention based on semantic/label graph, terme Model 4.

The following observations can be clearly found from Table~\ref{tab:tb4}:
1) Compared with Model 1 (baseline), the performance of Model 2 is improved on four datasets, which indicates the effectiveness of modeling the sample correlation.
2) Compared with Model 2, Model 3 significantly improves the performance of baseline (Model 1) on four datasets, showing the effectiveness of our proposed BGFormer (w/o LCG) by reducing the impact of negative samples.
3) Compared with Model 3, Model 4 further improves the performance of BGFormer (w/o LCG), which indicates the effectiveness of dual structure-constrained self-attention (SSA).
In summary, the effectiveness and superiority of each component in our proposed model can be fully verified with  the experimental results.

% parameter analysise

\subsection{Parameter Analysis}
The following three parameters are very important for our model, i.e., number of BGFormer layers $N$, $\alpha$ in Equ.(\ref{equ:eq12}) and $\lambda$ in Equ.(\ref{equ:eq7}). Several experiments are conducted on the CUB~\cite{welinder2010caltech} dataset to check their influence. As shown in Figure~\ref{fig:parm_Analysis} (b), when increasing the number of BGFormer layers $N$, we can find that the model performance increases until $N = 2$.
For parameter $\alpha$, as shown in Figure~\ref{fig:parm_Analysis} (c), the model performance is continuously improved until $\alpha = 0.6$.
For parameter $\lambda$, the performance is continuously improved until $\lambda = 0.4$, as illustrated in Figure~\ref{fig:parm_Analysis} (d). As a result, we set $N=2$, $\alpha=0.6$, and $\lambda=0.4$ in our experiments to obtain better results.

\section{Conclusion}
In this paper, we
rethink to explore sample relationships within each mini-batch for discriminative and robust data representation and learning.
To this end,
we first introduce a novel batch graph to jointly encode the visual and label relationships of samples via a unified model.
Then, we develop a novel BGFormer to learn sample-level representation.
BGFormer adopts the dual SSA mechanism to conduct message passing on the batch graph and fuses both visual and label cues together, followed by some FFN, LN operations, to obtain the enrichment data representations.
To validate the effectiveness of the proposed BGFormer, we apply it to metric learning tasks and obtain better performance on four benchmark datasets.

% \textbf{Limitations.}
% The current BGFormer has limited performance improvements for large-scale dataset with a large number of classes and a little samples per class.

%%%%%%%%% REFERENCES
{\small
\bibliographystyle{ieee_fullname}
\bibliography{reference}
}

\end{document}